\documentclass{article}

\usepackage{arxiv}
\usepackage{multirow}
\usepackage{amsmath}
\usepackage{subcaption}
\usepackage{xcolor}
\newcommand{\best}[1]{\textcolor{red}{\textbf{#1}}}
\newcommand{\secondbest}[1]{\textcolor{blue}{\textbf{#1}}}
\newcommand{\R}{\mathbb{R}}
\usepackage[utf8]{inputenc} % allow utf-8 input
\usepackage[T1]{fontenc}    % use 8-bit T1 fonts
\usepackage{hyperref}       % hyperlinks
\usepackage{url}            % simple URL typesetting
\usepackage{booktabs}       % professional-quality tables
\usepackage{amsfonts}       % blackboard math symbols
\usepackage{nicefrac}       % compact symbols for 1/2, etc.
\usepackage{microtype}      % microtypography
\usepackage{cleveref}       % smart cross-referencing
\usepackage{lipsum}         % Can be removed after putting your text content
\usepackage{graphicx}
\usepackage{natbib}
\usepackage{doi}

\title{Filter then Attend: Improving attention-based Time Series Forecasting with Spectral Filtering}

\newif\ifuniqueAffiliation
\uniqueAffiliationtrue

\ifuniqueAffiliation % Standard variant of author block
\author{Elisha Dayag\thanks{Use footnote for providing further
		information about author (webpage, alternative
		address)---\emph{not} for acknowledging funding agencies.} \\
	Department of Mathematics\\
	University of California, Irvine\\
	Irvine, CA 92617 \\
	\texttt{edayag@uci.edu} \\
	\And
	\hspace{1mm}Nhat Thanh Van Tran \\
	Department of Mathematics\\
	University of California, Irvine\\
	Irvine, CA 92617 \\
	\texttt{nhatt@uci.edu} \\
	 \AND
	 Jack Xin \\
     Department of Mathematics \\
	 University of California, Irvine \\
	 Irvine, CA 92617 \\
	 \texttt{jxin@math.uci.edu} \\
	 }
\else
\usepackage{authblk}

\fi

\hypersetup{
pdftitle={A template for the arxiv style},
pdfsubject={q-bio.NC, q-bio.QM},
pdfauthor={David S.~Hippocampus, Elias D.~Striatum},
pdfkeywords={First keyword, Second keyword, More},
}

\begin{document}
\maketitle

\begin{abstract}
	%\lipsum[1]
    Transformer-based models are at the forefront in long time-series forecasting (LTSF). While in many cases, these models are able to achieve state of the art results, they suffer from a bias toward low-frequencies in the data and high computational and memory requirements. Recent work has established that learnable frequency filters can be an integral part of a deep forecasting model by enhancing the model's spectral utilization. These works choose to use a multilayer perceptron to process their filtered signals and thus do not solve the issues found with transformer-based models. In this paper, we establish that adding a filter to the beginning of transformer-based models enhances their performance in long time-series forecasting. We add learnable filters, which only add an additional $\approx 1000$ parameters to several transformer-based models and observe in multiple instances 5-10 \% relative improvement in forecasting performance. Additionally, we find that with filters added, we are able to decrease the embedding dimension of our models, resulting in transformer-based architectures that are both smaller and more effective than their non-filtering base models. We also conduct synthetic experiments to analyze how the filters enable Transformer-based models to better utilize the full spectrum for forecasting.
\end{abstract}

% keywords can be removed
\keywords{Machine Learning \and Time Series Forecasting }

\section{Introduction}
Time series forecasting is a critical task in domains such as energy usage \cite{maleki2024future}, traffic prediction \cite{van2022comparison} and financial markets \cite{gajamannage2023real}. Given the resounding success of deep learning in the adjacent fields of computer vision and natural language processing, there has been much interest in applying deep learning to time series forecasting. While some applications of deep learning to the forecasting problem utilized architectures such as the LSTM \cite{kong2025unlocking}, the GRU \cite{zhang2024novel} and temporal convolution networks, over time two broad approaches to LTSF with deep learning have coalesced: multilayer perceptrons (MLP) and transformer-based methods. Naive application of the  transformer for forecasting have been deemed ineffective \cite{zeng2023transformers} in comparison to even linear models; adapting the transformer to be amenable to time series forecasting has required a merger with the techniques of classical signal processing and forecasting. Many of these mergers incorporate information from the frequency domain to help in forecasting \cite{yi2023survey}. This is because in the frequency domain, periodicities in the data become more apparent and are easier for the model to extract. In this paper, we present a simple merger of transformer-based models with the well-established signal processing technique of frequency filtering. Building upon recent work in LTSF that learned filters are effective for time series forecasting \cite{yi2024filternet} and recent work in computer vision that has indicated that learned filters can enhance the performance of vision transformers \cite{patro2025spectformer}, we demonstrate that combining frequency filters with attention-based transformers results in a model that is stronger than the sum of its parts. As proof of our findings, we present the simple baselines of FilterFormers obtained by passing our time series through a learnable filter before applying standard variants of the transformer. By filtering first, we are able to enhance the forecasting performance of our models, especially on larger datasets for longer prediction horizons, with little added computational cost (in many cases, we are able to speedup the model because we no longer need as large an embedding dimension to obtain similar or better results). 

Our contributions can be summarized as follows: 
\begin{itemize}
    \item Inspired by the recent successes of learnable filters in time series forecasting and computer vision, we combine a simple learnable filter with a transformer-based architecture to create FilterFormer. 
    \item We conduct experiments on a variety of datasets spanning different domains and characteristics and demonstrate that adding a learnable filter improves  performance for multiple types of transformer-based forecasters. 
    \item Via synthetic experiments, we provide evidence that filters improve the performance of transformer-based models by helping them capture high-frequency patterns in the data, countering the low-pass nature of transformers.
\end{itemize}

\section{Related Work and Motivations}
\subsection{Transformer-Based Architectures}
In recent years there are a variety of transformer based approaches for forecasting that we summarize here. Initial works attempted to replace vanilla attention to overcome the quadratic sequence complexity. Informer \cite{zhou2021informer} uses a ProbSparse self-attention and self-attention distilling to efficiently capture the most important keys. Autoformer \cite{wu2021autoformer} extracts seasonal and trend components of the time series and uses autocorrelation in place of self-attention. An individual time step may not possess significant semantic information, so some methods find innovation in applying the attention mechanism to things other than the pointwise time sequence.  \cite{nie2022time} segments time-series into subseries-level patches which are used as the tokens for transformer. iTransformer \cite{liu2023itransformer} embeds the raw time series of individual variates as tokens then performs self-attention to obtain multivariate correlations. Leddam  \cite{yu2024revitalizing} utilizes a dual attention module consisting of channel-wise self attention and autoregressive self-attention as well as a learnable convolution kernel.

\subsection{Frequency Learning in Time Series Forecasting}

Spectral analysis has a long history in time series forecasting  \cite{shumway2000time}. This is because time series data often contains periodic patterns which are reflected in the spectrum. Recent work has tried to combine spectral analysis with deep learning for improved forecasting. 
FEDFormer \cite{zhou2022fedformer} uses an Encoder-Decoder architecture both incorporating spectral analysis. The encoder and decoder both use the Discrete Fourier Transform (DFT) to transform the time signal into the frequency domain. They then randomly samples modes for computational efficiency. The encoder applies a parametrized kernel to the frequency representation to learn the data's temporal patterns before the inverse discrete Fourier transform (IDFT) is applied to obtain a time signal. The decoder applies self-attention in the frequency domain. 

Fredformer \cite{piao2024fredformer} applies patching and channel wise attention in the frequency domain combined with frequency normalization techniques to force the transformer encoder to utilize low energy frequency features in the data. This is done with the intent of mitigating frequency bias in the transformer. 

FreTS \cite{FReTS} transforms the time signal into the frequency domain using the  DFT. A multilayer perceptron (MLP) is applied in the frequency domain before. FreTS also applies a similar process along the channel dimensions to obtain cross-channel dependencies.  

FITS \cite{xu2024fits} explores the ability of frequency learning to develop a compact baseline for time series forecasting. In this work the authors take the DFT of the signal, apply a low pass filter to it, then applies a learnable (complex) linear mapping in the frequency domain before recovering the signal using the IDFT. 

Part of the innovation of FITS is applying a frequency filter to the data before the learnable mappings. Rather than using a handcrafted filter, FilterNet  \cite{yi2024filternet} randomly initializes a learnable filter then performs multiplication with the input signal in the frequency domain. After converting back to a time signal using the IDFT, the filtered signal is processed using a MLP.

We believe that these works have demonstrated the efficacy of frequency learning as a technique in time series forecasting. Unlike some of these methods, we believe that the power of frequency filters is best utilized as a building block enabling the transformer. 

%\subsection{Channel Dependency Strategies in Multivariate Time Series Forecasting} A question of interest in LTSF is how to model dependencies between the different features in a time series. The two modeling philosophies can be summarized as Channel Independent (CI) and Channel Mixing (CM). CI methods model each feature seperately, ignoring any cross-channel correlations. Channel independence has been shown to dramatically improve forecasting performance with regards to electricity demand and traffic flow. By learning separate attention weights and a separate linear head for each feature, models are robust to individual series having  unique behaviors. This attention to specificity outweighs the potential of cross-channel information when the inter-channel dependencies are weak or noisy. In addition, it is theorized that channel independence prevents models from overfitting and allows them to converge faster than models that utilize channel mixing. 

%Conversely, channel dependent strategies 

\section{Preliminaries}

The discrete Fourier transform converts a signal $x \in \mathbb{R}^N$ into its frequency domain representation $\hat{x} \in \mathbb{R}^N$
\begin{align}
\hat{x}_n = \sum_{k=0}^{N-1} x_n e^{\frac{-i2\pi nk}{N} }
\end{align}
The DFT is Hermitian, can be performed in $O(N\log N)$ operations using the fast Fourier transform  \cite{cooley1965algorithm}, and is reversible; once a signal has been processed in the frequency domain, we can convert it back to the time domain using the Inverse Discrete Fourier Transform (IDFT) 
\begin{align}
    x_n = \sum_{k=0}^{N-1} \hat{x}_k e^{\frac{i2\pi kn}{N}}
\end{align}

 According to the convolution theorem \cite{proakis2007digital}, the Fourier transform of the a convolution of two signals is equal to the pointwise product of their respective Fourier transforms in the frequency domain. So, we can use the DFT to create frequency filters that modify the spectral content of our signals. 

 A frequency filter is any system used to modify the frequency components of a signal. In this paper we consider a frequency filter $P$ to be the Fourier transform of a fixed signal $p \in \mathbb{R}^n$. $P$ is applied to our signal $\hat{x}$ via the pointwise product to produce an output signal $\hat{y}$ with frequency representation 
\begin{equation}\label{frequency filter}
    \hat{y}_k = P_k \cdot \hat{x}_k 
\end{equation}
so that $$y  =p * x,$$ where $p$ is the inverse DFT of $P$ and $*$ is circular convolution. It is the norm in digital signal processing to use hand-crafted filters. In computer vision \cite{patro2025spectformer} as well as in time series forecasting with FilterNet,  data-driven learnable filters have also been shown to be effective for tasks like classification and forecasting. In the former, it was shown that using  learned frequency filters before applying a vision transformer (i.e. self-attention applied to patches) outperformed attention-only and spectral-mixing only approaches on image classification. We show that a similar truth holds for time series forecasting.

\section{Methods}
In our experiments, we are given a historical time series with lookback length $L$, where each timestamp has $D$ variates: $X = \{X_1, \dots, X_L\} \in \R^{D\times L}$. Our problem is to predict the next $H$ timesteps, denoted by $\{X_{L+1} ,\dots, X_{L+H}\} \in \R^{D\times H}$.

\subsection{Architecture}
 In our work, we also attach learnable filters to the architectures of PatchTST, iTransformer and Leddam giving us FilterFormer, iFilterFormer, and FilterLeddam respectively. For the purposes of having a simpler model to ablate with and examine, we detail the construction of FilterFormer and use it for many of our experiments. As a demonstration of the combined power of filtering and attention, we present a simple model combining both spectral and attention blocks to learn periodic and aperiodic representations of our signal.

\begin{figure}[h]
    \centering
    
        \includegraphics[width=\linewidth]{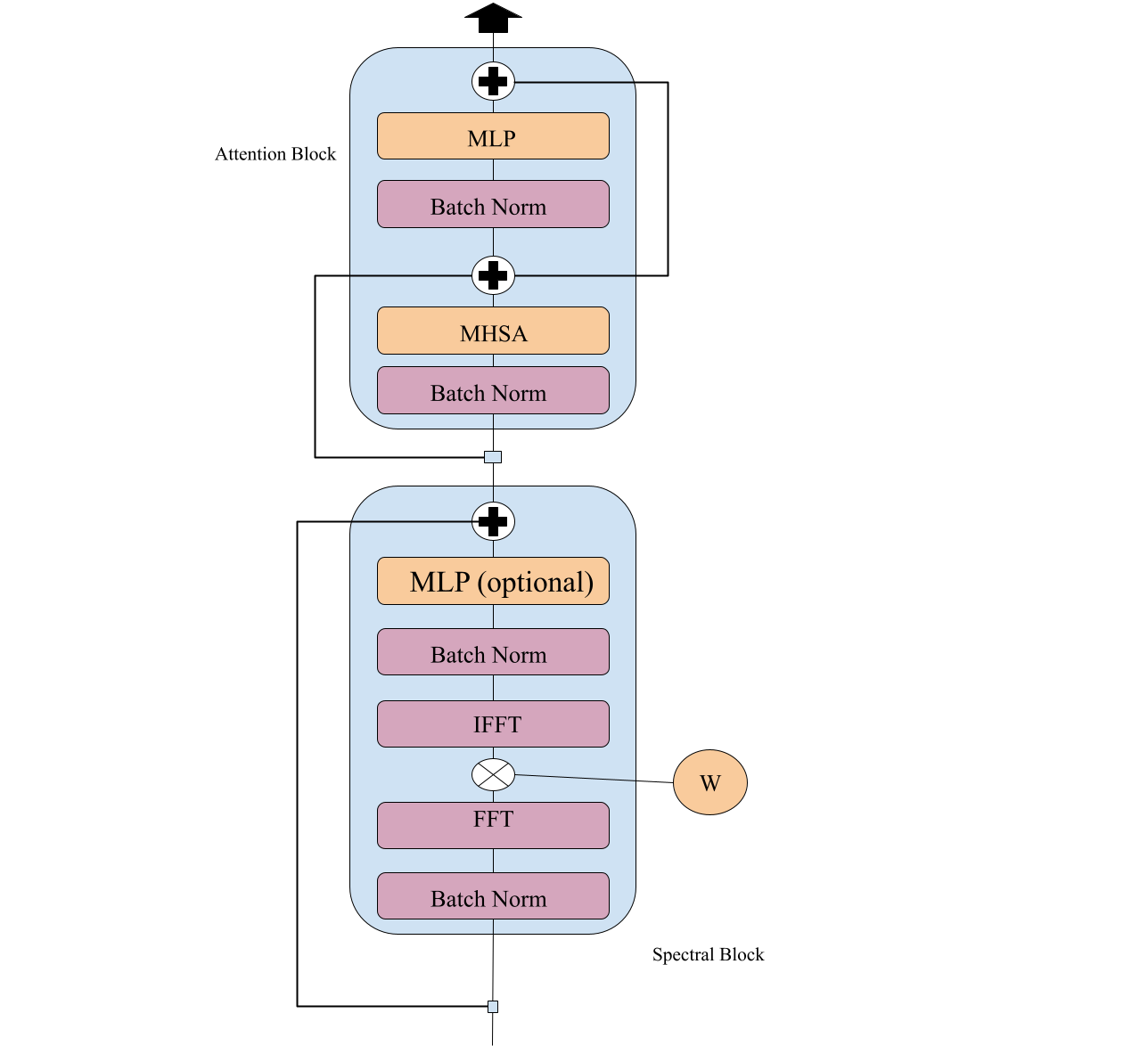}

    \caption{The designs of our spectral block and our attention block. }
    \label{fig:enter-label}
\end{figure}

\subsubsection{Input Embedding}
We start by converting our signal $X$ into a series of patches.  Following the framework of PatchTST, we treat our signal as $D$ independent univariate time series. We denote the ith series as $X^{(i)} \in \mathbb{R}^L$. $X^{(i)}$ is normalized to have zero mean and unit variance via reversible instance normalization \cite{kim2021reversible}. We then divide $X^{(i)}$ into a sequence of consecutive overlapping or non-overlapping patches. Given a patch size $P$, the total number of patches is $$N = \frac{\lfloor L - P\rfloor}{P} + 1.$$ By turning our sequence into patches we are able to reduce our computational burden, allowing us to work with full self-attention later on, and we can aggregate local information into each patch, giving each spot into our self-attention more semantic meaning. After patchifying our sequence, we embed each patch $X_p^{(i)}\in \mathbb{R}^{P\times L_P}$ into a latent space representation $Y_p^{(i)} \in \mathbb{R}^{P \times d_m}$ using a learnable linear layer. We enhance each latent space representation by adding a positional embedding.
\subsubsection{Spectral Block}
The goal of the spectral block is to capture the different frequency components of the sequence to help model temporal dependencies in the data. After performing batch normalization \cite{ioffe2015batch} we pass the signal through our spectral gating network. For a given embedded patch $Y^{(i)}$ (denoted as $Y$ from hereon) and its frequency domain representation $\hat{Y}$, our spectral gating in the frequency domain can be expressed as a global convolution in the time domain. To realize our filter, we take the rFFT of an initially random set of real parameters which we denote by $W$. We then perform pointwise multiplication in the frequency domain before reverting our filtered signal back into the time domain using the IDFT.  As an aside, because both our signal and the time representation of our filter are real-valued, we can be assured that their product is also the DFT of a real-valued 
\begin{figure}
    \centering
    \includegraphics[width=0.5\textwidth,height=7cm]{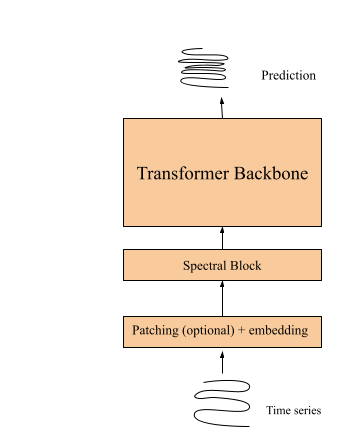}
    \caption{The general form of all the FilterFormer models created in our paper. }
    \label{fig:placeholder}
\end{figure}
signal given that it has the same conjugate symmetry: 
\begin{align*}
    \overline{W[k]Y[k]} &= \overline{W[k]} \cdot \overline{Y[k]} \\
    &= W[-k]\cdot Y[-k].
\end{align*} Thus, when we take the IDFT of our product, we are sure to obtain a real-valued signal again. The total operation of our spectral block can be written as: 
\begin{align}
    \mathcal{F}^{-1} (\mathcal{F}(W) \cdot  \mathcal{F}(Y))= \mathcal{F}^{-1}(\mathcal{F}(W*Y))
\end{align}
where * denotes circular convolution, and $\mathcal{F}$ denotes the DFT. After filtering our signal using the spectral gating network, we perform instance normalization once again before passing it through an MLP to further extract a frequency representation of our signal. We note that our spectral block is similar to FilterNet. The primary differences are that 1) we are passing an embedded version of patches of our signal through our filters and 2) because we are passing our filtered signal into a transformer, we find it is not necessary to include a MLP in our spectral block as FilterNet does. While FilterNet should be lauded for its ability to extract periodic trends in the data using a relatively lightweight scheme, we believe that better forecasting can be achieved by passing the filtered signals through a self-attention mechanism. We also note that recent work has adopted more complicated mechanisms inside of the filter block \cite{eldele2024tslanet}, but that is outside of the scope of this paper, as we are trying to demonstrate that filtering, even in its simplest form, can improve transformed-based models. 
\subsubsection{Attention Block}
In the attention block, we opt to go with standard multi-head attention on our filtered patches. Each head $h = 1, \dots, H$ in multi-head attention transforms our patch $Y_p^{(i)}$ into query matrices $Q_h^{(i)} = (Y_p^{(i)})^TW_h^Q$,  key matrices $K_h^{(i)} = (Y_p^{(i)})^TW_h^K$, and value matrices $V_h^{(i)} = (Y_p^{(i)})^TW_h^V$, where $W_h^Q,W_h^K \in \mathbb{R}^{d_m \times d_k}$ and $W_h^V \in \mathbb{R}^{d_m \times d_m}$. After that we use the scaled dot-product to get our attention output $O_h^{(i)} \in \mathbb{R}^{d_k\times N}$: $$(O_h^{(i)})^T = \text{Softmax}\left(\frac{Q_h^{(i)} K_h^{(i)^T}}{\sqrt{d_k}} \right) V_h^{(i)}$$
Our multi-head attention block also contains batch normalization and a multilayer perceptron with residual connections to process our attention output. Afterwards, it generates the representation $z^{(i)} \in \mathbb{R}^{d_k\times N}$. Finally we use a flatten layer with a linear head to map our representation to the prediction result. Like PatchTST we choose to share our weights across each channel. This enables our model to stay light.

 Our model has $L$ layers in total, $\alpha$ of which are spectral blocks and $L-\alpha$ of which are attention blocks. $\alpha$ is a hyperparameter allowing us to decide the tradeoff between attention and spectral layers appropriate for the dataset in question. In general, for datasets with a smaller number of channels/features, we opt for a smaller number of layers to prevent overfitting. Given the nature of a frequency filter, one may question the need for multiple spectral blocks. We find that when incorporating the MLP in the spectral block, there are situations where having multiple spectral blocks is beneficial for forecasting performance (see appendix).

\section{Experiments}
We evaluate our model on nine popular datasets, namely the four ETT datasets (ETTm1, ETTm2, ETTh1, and ETTh2), Exchange-rate, Weather, Electricity, Solar-Energy, and Traffic. We preprocess our datasets using standard normalization methods. The ETT (Electricity Transformer Temperature) dataset contain data collected from electricity transformers every 15 minutes (in the case of ETTm1 and ETTm2) or 1 hour (in the case of ETTh1 and ETTh2) between July 2016 and July 2018. The exchange dataset contains the daily exchange rates of  Singapore, Australia, British, Canada, Switzerland, China, Japan, and New Zealand ranging from 1990 to 2016. The Weather dataset 
contains 21 meteorological indicators including air temperature, atmospheric pressure, and humdity. The data was collected every 10
minutes from the Weather Station of the Max Planck Institute for Biogeochemistry in 2020. The Traffic  dataset contains hourly traffic data measured by 862 sensors on San Francisco Bay area
freeways. The data, which describes lane occupancy rates (between 0 and 1) was  collected from 2015 to 2017. The Electricity dataset collects the
hourly electricity consumption of 321 clients from 2012 to 2014. The Solar-energy dataset contains the solar power production records in 2006, sampled every ten minutes from 137 PV plants in the state of Alabama.

\begin{table}[h]
\centering
\setlength{\tabcolsep}{1mm}
\scalebox{0.50}{
\begin{tabular}{lccccccccc}
\toprule
\textbf{Datasets} & \textbf{ETTh1} & \textbf{ETTh2} & \textbf{ETTm1} & \textbf{ETTm2} & \textbf{Electricity} & \textbf{Weather} & \textbf{Traffic} & \textbf{Exchange} & \textbf{Solar-Energy} \\
\midrule
Variables & 7 & 7 & 7 & 7 & 321 & 21 & 862 & 8 & 137 \\
Timesteps & 17420 & 17420 & 69680 & 69680 & 26304 & 52696 & 17544 & 7588 & 52560 \\
Frequency & Hourly & Hourly & 15min & 15min & Hourly & 10min & Hourly & Daily & 10min \\
Information & Electricity & Electricity & Electricity & Electricity & Electricity & Weather & Traffic & Economy & Energy \\
\bottomrule

\end{tabular}}
\caption{The details of the datasets used in our experiments.}
\end{table}

\subsubsection{Baselines}
We compare our model to iTransformer \cite{liu2023itransformer}, patchTST \cite{nie2022time}, and Leddam  \cite{yu2024revitalizing} the models used as a base to construct our FilterFormers. We also compare our model to other frequency based methods like  FilterNet and Fredformer  as well as the baseline MLP method DLinear. For FilterNet, we are explicitly comparing to their PaiFilter, as this learnable filter is most similar in design to the one used in Filterformer. For each model we fix the lookback length to $96$, and present the multivariate forecasting results for the prediction horizons $H \in \{96,192,336,720\}$.

\subsubsection{Implementation Details} Our model is implemented with Pytorch 1.12 and all experiments were conducted on a cluster of four NVIDIA RTX 2080 Ti's. We use mean square error (MSE) as our loss function and report MAE (mean absolute error) and MSE as our evaluation metrics. For baselines, we either use their reported numbers as stated or run the relevant training scripts found on their respective codebases. FFor FilterFormer, we experimented with an embedding dimension of 128 and 256. For iFilterFormer and FilterLeddam, we performed runs with embedding dimension 128, 256, and 512. For FilterLeddam , the embedding dimension usually matched that which the authors of the base model \cite{yu2024revitalizing} specified in their uploaded code. For iFilterFormer, we generally opted for smaller embedding dimensions than the authors of iTransformer \cite{liu2023itransformer} used in the interest of efficiency. We trained our models using the Adam \cite{kingma2014adam} optimizer for 50 epochs with an early stopping procedure if validation loss had not increased in 15 epochs (we found this early stopping length necessary for training on the solar-energy dataset, where validation loss may be unchanged for 11 epochs before decreasing). For FilterLeddam and iFilterFormer we primarily matched our conditions like learning rate, dropout, and number of attention layers to those of the base model paper. Additionally, for these models, we did not include a multilayer perceptron in the spectral block. For FilterFormer, we experimented with learning rates in $\{1e-4, 5e-4, 1e-3\}$. For FilterFormer, we primarily performed runs using three attention blocks and either 1 or 2 spectral blocks. The spectral blocks for FilterFormer did include a multilayer perceptron.  

\begin{table*}[t]
\centering
\setlength{\tabcolsep}{1mm}
\caption{Model Performance Comparison (MAE/MSE). Lower is better. Best and second best values in each row colored in red or blue, respectively. }
\scalebox{0.70}{
\begin{tabular}{ll|cc|cc|cc|cc|cc|cc|cc|cc|cc}
\hline
\multicolumn{2}{c|}{} & \multicolumn{2}{c|}{FilterFormer} & \multicolumn{2}{c|}{FilterNet} & \multicolumn{2}{c|}{iTransformer} & \multicolumn{2}{c|}{PatchTST} & \multicolumn{2}{c|}{Leddam} & \multicolumn{2}{c|}{iFilterFormer} & \multicolumn{2}{c|}{DLinear} & \multicolumn{2}{c|}{Fredformer} & \multicolumn{2}{c}{FilterLeddam} \\
Models & Metrics & MSE & MAE & MSE & MAE & MSE & MAE & MSE & MAE & MSE & MAE & MSE & MAE & MSE & MAE & MSE & MAE & MSE & MAE \\
\hline
\multirow{4}{*}{ETTm1}
 & 96 & \best{.318} & \best{.354}   & \best{.318} & \secondbest{.358} & .334 & .368 & .329 & .367 & .329 & .361 & .342 & .373 & .344 & .370 & .326 & .361 & .327 & .361 \\
 & 192 & \best{.361} & .384 &  .364 & .383 & .377 & .391 & .367 & .385 & .369 & .383 & .379 & .392 & .379 & .393 & \secondbest{.363} & \best{.380} & .364 & \secondbest{.381} \\ % Corrected: ETTm1 192 MAE TexFilter .336 (R), PatchTST .367(B). MSE SC .3641(R), Tex/Times .387(B)
 & 336 & \best{.385} & \secondbest{.403}   & .396 & .406 & .426 & .420 & .399 & .410 & .394 & \best{.402} & .415 & .411 & .410 & .411 & .395 & .403 & \secondbest{.393} & .404 \\
 & 720 & \best{.450} & \best{.438} & .456 & .444 & .491 & .459 & .454 & \secondbest{.439} & .460 & .442 & .486 & .450 & .473 & .450 & \secondbest{.453} & \best{.438} & .462 & .440 \\ % Corrected: ETTm1 720 MAE TexFilter .447 (R), SC .4496 (B). Image has .447 as bolded. If I read .447 for SC then SC is red. User prompt for SC is .4496. So Tex .447(R), SC .4496(B). PatchTST MSE .439(R), Tex/iTr/FITS/DLinear/RLinear/TimesNet (multiple choices for blue if .448/.449/.450 are close). TexFilter .448 (B).
\hline
\multirow{4}{*}{ETTm2}

 & 96  & .176 & .261   & \best{.174} & \best{.257} & .180 & .264 & \secondbest{.175} & \secondbest{.259} & .176 & \best{.257} & .185 & .267 & .187 & .281 & .177 & \secondbest{.259} & .179 & .261 \\
 & 192 & \best{.235} & \secondbest{.301}   & \secondbest{.240} & \best{.300} & .250 & .309 & .241 & .302 & .243 & .303 & .243 & .309 & .272 & .349 & .243 & \secondbest{.301} & \secondbest{.240} & .303 \\
 & 336 &  \best{.295} & \best{.339}  & .297 & \best{.339} & .311 & .348 & .305 & .343 & .303 & .341 & .310 & .351 & .316 & .372 & .302 & \secondbest{.340} & \secondbest{.296}& \secondbest{.340} \\
 & 720 & \secondbest{.394} & .394 & \best{.392} & \secondbest{.393} & .412 & .407 & .402 & .400 & .400 & .398 & .403 & .403 & .452 & .457 & .397 & .396 &  \best{.392} & \best{.391} \\
\hline
\multirow{4}{*}{ETTh1}
 & 96  & \best{.373} & .395  & \secondbest{.375} & \secondbest{.394} & .386 & .405 & .414 & .419 & .377 & .394 & .390 & .404 & .383 & .396 & \best{.373}  & \best{.392} & .383 & .397 \\
 & 192 & \secondbest{.425} & .431  & .436 & \secondbest{.422} & .441 & .436 & .460 & .445 & \best{.424} & \secondbest{.422} & .441 & .437 &  .433 & .426 & .433 & \best{.420}& .433 & 
 .428 \\
 & 336 & \secondbest{.460} & .454  & .476 & \secondbest{.443} & .487 & .458 & .501 & .466 & \best{.459} & \best{.442} & .480 & .457 & .479 & .457 & .470 & .437 & .464 & .444 \\
 & 720  &  .469 & .477  & .474 & .469 & .491 & .497 & .500 & .488 & \best{.463} & \best{.459} & .471 & .472 & .517 & .513 & .467 & \best{.459} & \secondbest{.465} & 
 \secondbest{.464} \\ % Corrected ETTh1 720 MAE: RLinear .470 (R), SC .4809 (B). MSE RLinear .470 (R), SC .4792 (B)
\hline
\multirow{4}{*}{ETTh2}
 & 96 & \best{.292} & .344  & \best{.292} & \secondbest{.343} & .297 & .349 & .302 & .348 & \best{.292} & .344 & .306 & .351 & .320 & .374 & .293 & \best{.342} & .295 & .350 \\
 & 192 & \best{.364} & .391  & .369 & .395 & .380 & .400 & .388 & .400 & \secondbest{.367} & \best{.389} & .374 & .398 & .449 & .454 & .371 & \best{.389} & .368 & \best{.389} \\
 & 336 & .416 & .426 & .420 & .432 & .428 & .432 & .426 & .433 & \secondbest{.412} & .424 & .416 & .452 & .467 & .469 & \best{.382} & \best{.409} & \secondbest{.412} & \secondbest{.422} \\
 & 720 & .424 & .445  & .430 & .446 & .427 & .445 & .431 & .446 & .419 & .438 & .420 & .442 & .656 & .571 & \secondbest{.415} & \secondbest{.434} &.\best{412} & \best{.433} \\
\hline
\multirow{4}{*}{ECL}
 & 96  & .154 & .246 & .176 & .264 & .148 & .240 & .181 & .270 & \best{.141} & \best{.235} & .148 & .241 & .195 & .277 & .147 & .241 & \best{.141} & \secondbest{.239} \\
 & 192 & .165 & .256 & .185 & .270 & .162 & .253 & .188 & .274 & \secondbest{.159} & \secondbest{.252} & .163 & .289 & .194 & .280 & .165 & .258 & \best{.157} & \best{.250} \\
 & 336 & .183& .275 &.202 & .286 & .178 & .269 & .204 & .293 & \best{.173} & \secondbest{.268} & .176 & .300 & .207 & .296 & .177 & .273 & \secondbest{.174} & \best{.260} \\
 & 720 & .221 & .310 & .242 & .319 & .225 & .317 & .246 & .324 & \secondbest{.201} & \secondbest{.295} & .208 & .305 & .242 & .329 & .208 & .304 & \best{.200} & \best{.294} \\
\hline
\multirow{4}{*}{Exchange}
 & 96  & .085 & \best{.201} & \best{.083} & \secondbest{.202} & .086 & .206 & .088 & .205 & .086 & .207 & .088 & .209 & .085 & .210 & .084 & .207 & \secondbest{.084} & .203 \\
 & 192 & .186 & .303 & \best{.174} & \secondbest{.296} & .177 & \secondbest{.296} & .176 & .299 & \secondbest{.175} & .301 & .180 & .302 & .178 & .299 & .178 & .392 & \secondbest{.175} & \best{.295} \\
 & 336 &  .329 & .416 & .326 & .413 & .331 & .417 & \secondbest{.301} & \best{.397} & .325 & .415 & .314 & .407 & \best{.298} & .409 & .319 & .408 & .306 & \secondbest{.400} \\
 & 720 & .835 & .681 & .840 & .670 & .847 & .691 & .901 & .714 & .831 & \best{.636} & \best{.726} & \secondbest{.639} & .861 & \textbf{.671} & \secondbest{.749} & .651 & .801 & .667 \\ % Exchange 720 MAE: PatchTST .691 (R), FITS .828 (B). MSE: iTransformer .647 (R), DLinear .671 (B)
\hline
\multirow{4}{*}{Traffic}
 & 96  & .470 & .293 & .506 & .336 & \best{.395} & \best{.268} & .462 & .295 & .426 & .376 & \secondbest{.396} & \secondbest{.270} & .650 & .397 & .406 & .277 & .419 & .277 \\
 & 192 & .482 & .298 & .508 & .333 & \best{.417} & \best{.276} & .466 & .296 & .458 & .289 & \secondbest{.418} & \secondbest{.277} &.600 & .372 & .426 & .290 & .435 & .288  \\
 & 336 & .504 & .301 & .518 & .335 & \secondbest{.433}&\secondbest{.283} & .482 & .304 & .486 & .297 & \best{.425} & \best{.280} & .606 & .374 & .437 & .292 & .451 & .297 \\
 & 720 & .528 & .312 & .553 & .354 & \secondbest{.467} & \secondbest{.302} & .514 & .322 & .498 & .313 & \best{.458} & \best{.300} & .646 & .395 & .462 & .305 & .490 & .314  \\
\hline
\multirow{4}{*}{Weather}
 & 96  & \secondbest{.161} & .205 & .164 & .210 & .174 & .214 & .177 & .218 & \best{.156} & \secondbest{.202} & .170 & .220 & .194 & .248 & .163 & .207 & \best{.156} & \best{.201} \\
 & 192 & .208 & \best{.248} & .214 & .252 & .221 & .254 & .225 & .250 & \secondbest{.207} & \secondbest{.250} & .222 & .261 & .234 & .290 & .211 & .251 & \best{.206} & \best{.248} \\
 & 336 & .263 & \best{.288} & .268 & .293 & .278 & .296 & .278 & .297 & \secondbest{.262} & .291 & .276 & .306 & .283 & .335 & .267 & .292 & \best{.260} & \secondbest{.289} \\
 & 720 & \secondbest{.344} & \best{.338} & \secondbest{.344} & .342 & .358 & .347& .354 & .348 & \best{.343} & .343 & .358 & .352 & .348 & .385 & \best{.343} & \secondbest{. 341} & .345 & \secondbest{.341} \\

 \hline 
 \multirow{4}{*}{Solar-Energy}& 96  & .199 & .243 & .283 & .293 & .203 & .237 & .234 & .286 & \secondbest{.197} & \secondbest{.241} & .198 & .232 & .290 & .348 & \best{.185} & \best{.233} & \secondbest{.197} & \secondbest{.241} \\
 & 192 & .238 & .281 & .321 & .311 & .233 & \secondbest{.261} & .267 & .310 & .231 & .264 & .232 & \secondbest{.261} & .320 & .398 & \best{.227} & \best{.253} & \best{.227} & .264 \\
 & 336 & .258 & .288 & .369 & .334 & .248 & \secondbest{.273} & .290 & .315 & \best{.241} & \best{.268} & .248 & .278 & .353 & .415 & \secondbest{.246} & .284 & .247 & .277 \\
 & 720 & .260 & .291 & .373 & .336 & .252 & .286& .289 & .317 & .250 & .281 & .252 & .282 & .356 & .413 & \best{.247} & \best{.276} & \best{.247} & \best{.276} \\

\hline
\multicolumn{2}{l}{1st Counts} & 9& 7& 6& 3& 2& 2&0 &1 & 9 & 7 & 3 & 2 & 1 & 0 & 5 & 9 & 10 & 9 \\

\end{tabular}%
}
\label{tab:model_performance_best_scores}
\end{table*}

\subsubsection{Main Results}
\begin{figure}[h]
    \centering
    
    \includegraphics[width=.75\linewidth]{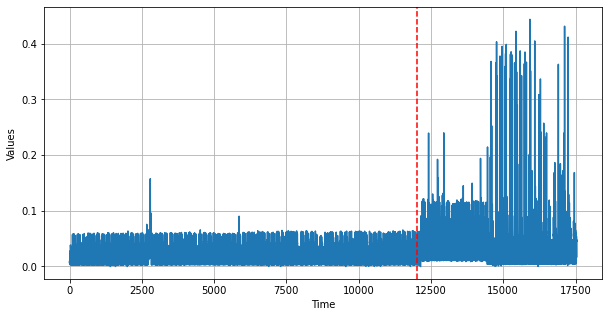}
    \caption{a) Channel 840 of the traffic dataset. The red line indicates the split between the training data and the validation/test data. }
    \label{traffic 840}
\end{figure}

From the table, we see that across a variety of datasets, our FilterFormers outperform most of our baseline models. The performance of the filtered model and the base model is highly correlated; filters are not a panacea and are ultimately limited by the forecasting power of the backbone they are attached to. We focus for a moment on our simplest model, FilterFormer. While it shows improvement over PaiFilter (FilterFormer with no attention) in general, this improvement is most apparent on the larger ECL and Solar-Energy datasets. This implies that on complex datasets, forecasting models with filters need a larger capacity than a simple MLP, which for FilterFormer comes in the form of multi-head self-attention. We note that on the traffic dataset, FilterFormer performs worse than its base model PatchTST. This occurs because some channels of the traffic dataset show a massive discrepancy between the train and test sets, leading to a large discrepancy in the Fourier spectra between testing and training (see Figure \ref{traffic 840}). 

When we remove even one of these outliers, we see the performance of FilterFormer dramatically improve. The results of our model and some of the baselines when trained on the traffic dataset with one single channel removed are presented in Table \ref{tab:traffic_test}. While every model exhibits better forecasting performance on this new dataset, our FilterFormer's performance is now competitive with all of our baselines and is significantly more powerful than PatchTST. In light of this result, we can see that for a complex dataset adding filtering can result in improved forecasting for a transformer-based model. This also corroborates the findings of \cite{yi2024filternet} that filtering alone is not optimal for larger, more complex datasets. 

The improvements of our three models over their not filtered counterparts can be shown in Table \ref{tab:model_comparison}. We see that on most datasets, adding filtering leads to improvements. While some of these improvements may seem negligible, we want to emphasize that 1) filtering imposes negligible added compute to a base model and 2) on many of these datasets, single digit improvements in relative forecasting performance are usually the result of dramatic changes in architecture and complexity. In this regard, we view filtering as a free lunch: it is a minor addition that rarely hurts performance, but it has the potential to help greatly. We spend the rest of the paper examining the structure and outputs of these filter models to better understand why they outperform their nonfiltering counterparts. 

\begin{table}[h]
    \centering
    \scalebox{.75}{
\begin{tabular}{ll|cc|cc|cc|cc|cc|cc|cc}
\hline
\multicolumn{2}{c|}{Models} & \multicolumn{2}{c|}{FilterFormer} & \multicolumn{2}{c|}{FilterNet} & \multicolumn{2}{c|}{iTransformer} & \multicolumn{2}{c|}{PatchTST}  & \multicolumn{2}{c|}{FilterLeddam} \\
 MSE & & .361 & & .449 & & .359 & & .401 && .399\\
\hline
\end{tabular}
}
    \caption{Performance of the baseline models and our FilterFormer on the traffic dataset after the removal of channel 840. For this experiment, the prediction horizon was 720.}
    \label{tab:traffic_test}
    
\end{table}

\begin{table}[h]
    \centering
    \scalebox{.9}{
    \begin{tabular}{|l||cc|cc|cc|}
        \hline
        \multirow{2}{*}{Dataset} & 
        \multicolumn{2}{c|}{FilterFormer} & 
        \multicolumn{2}{c|}{iFilterFormer} & 
        \multicolumn{2}{c|}{FilterLeddam} \\
        \cline{2-7}
        & MSE & $\Delta\%$ & MSE & $\Delta\%$ & MSE & $\Delta\%$ \\
        \hline\hline
        ETTm1        & 0.378 & 2.3 & 0.405 & 0.3 & 0.386 & 0.0 \\
        \hline
        ETTm2        & 0.275 & 4.8 & 0.285& 1.0 & 0.276 & 1.9 \\
        \hline
        ETTh1        & 0.428 & 8.5 & 0.452 & 1.2 & 0.436 & -1.5 \\
        \hline
        ETTh2        & 0.374 & 3.2 & 0.372 & 1.0 & 0.371 & 0.6 \\
        \hline
        ECL          & 0.180 & 11.7 & 0.173 & 2.5 & 0.168 & 0.5 \\
        \hline
        Exchange     & 0.358 & 2.1 & 0.327 & 9.2 & 0.341 & 4.0 \\
        \hline
        Traffic      & 0.496 & -3.1 & 0.424 & 0.9 & 0.448 & 4.1 \\
        \hline
        Weather      & 0.244 & 5.6 & 0.256 & 0.0 & 0.241 & 0.1 \\
        \hline
        Solar-Energy & 0.238 & 11.8 & 0.232 & 0.4 & 0.229 & 0.4 \\
        \hline
    \end{tabular}}
    \caption{Model comparisons showing average MSE over all prediction lengths and percentage improvement over baseline models. 
             $\Delta\%$ represents improvement over: PatchTST (FilterFormer), 
             iTransformer (iFilterFormer), and Leddam (FilterLeddam).}
    \label{tab:model_comparison}
\end{table}

\subsection{Model Analysis}

\subsubsection{Extending Lookback Length}
Prior work indicates that one advantage of PatchTST over other transformer based forecasters is its ability to improve the quality of predictions by increasing the lookback length \cite{zeng2023transformers} \cite{nie2022time}. We investigated whether adding spectral blocks still enhances forecasting ability when the lookback length has been increased. We tested our FilterFormer under the same conditions except for a lookback length of 336. 

% Requires: \usepackage{colortbl}
\begin{table}[h]
    \centering
    \scalebox{.9}{
    \begin{tabular}{|l||cc|cc|cc|}
        \hline
        \multirow{2}{*}{Dataset} & 
        \multicolumn{2}{c|}{FilterFormer} & 
        \multicolumn{2}{c|}{PatchTST} & 
        \multicolumn{2}{c|}{Improvement} \\
        \cline{2-7}
        & MSE & MAE & MSE & MAE & MSE & MAE \\
        \hline\hline
        ECL          & 0.161 & .255 & 0.162 & .254 & 0.1 & -0.4 \\
        \hline

        Traffic      & 0.404  & 0.270 & .448 & .277 & 9.8 & 2.8   \\
        \hline
        Weather      & 0.225 & 0.263 & 0.229 & 0.265 &1.8  & 0.8 \\ \hline
    \end{tabular}}
    \caption{Model comparisons showing average MSE over all prediction lengths and percentage improvement over PatchTST/42 when the lookback length is increased to 336.}
    \label{lookback_length}
\end{table}

\subsubsection{Visualization of Predictions}
In Figure \ref{fig:signals} we present some visualizations of our models on the electricity dataset.  
\begin{figure}
    \centering
    \begin{subfigure}[h]{0.45\textwidth}
        \centering
        \includegraphics[width=\linewidth]{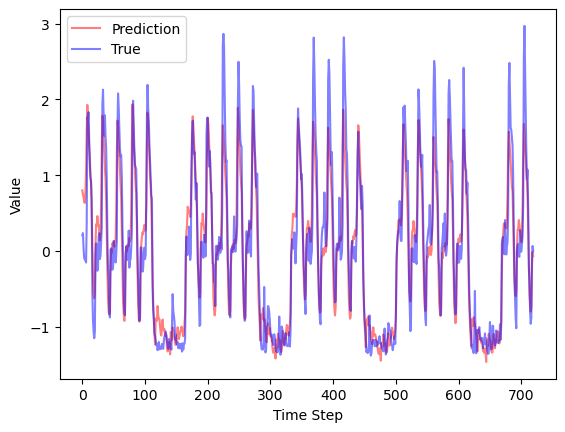} % Replace with your first image
        \caption{FilterFormer} % Fill in the MSE value
        \label{fig:filter_prediction1}
    \end{subfigure}
    \hfill
    \begin{subfigure}[h]{0.45\textwidth}
        \centering
        \includegraphics[width=\linewidth]{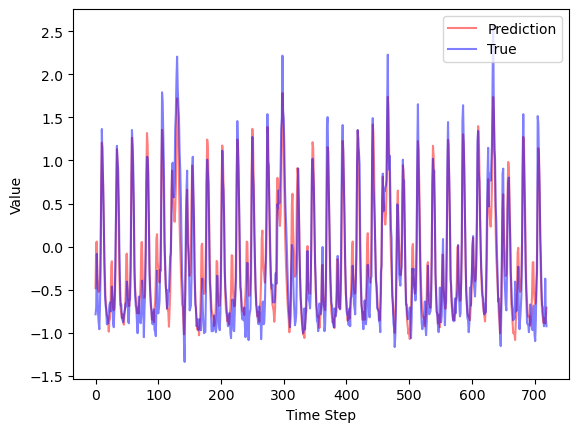} % Replace with your second image
        \caption{iFilterFormer} % Fill in the MSE value
        \label{fig:filterform_prediction}
    \end{subfigure}
    \caption{Visualizations of our models' predictions on the input 96, prediction 720 task of the ECL dataset.}

    \label{fig:signals}
\end{figure}
With a learned filter, these models are able to effectively capture high frequency patterns in the data over large prediction horizons, leading to the improvements in forecasting performance.

\subsubsection{Model Efficiency}
Here, let $d$ denote the embedding dimension of our model, $L$ the lookback length, and $H$ the prediction horizon. 

A single spectral block contains parameters from: 1) The batch normalization layers, 2) The actual spectral filter, and 3) the multilayer perceptron that processes the filtered signal before sending it to the next block. 
\begin{figure}[h]
    \centering
    \includegraphics[width=.75\linewidth]{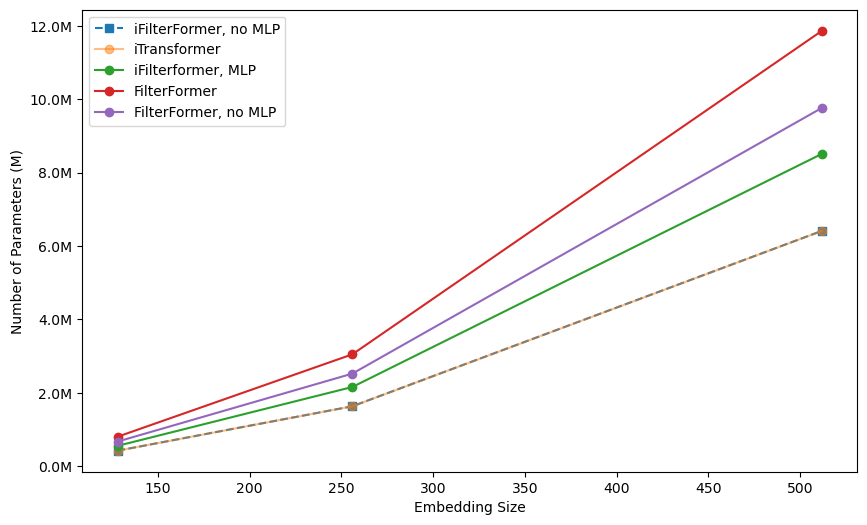}
    \caption{Number of parameters found in various implementations of our models with three attention layers, compared with iTransformer. Note that in practice the vast majority of our models have embedding dimension of 128 or 256. }
    \label{parameter list}
\end{figure} 
The batch layers and filter contribute about $\mathcal{O}(d)$ parameters each to our model, while the MLP contributes about $\mathcal{O}(Ld + Hd)$ parameters. We have that the majority of parameters in this block come from our multilayer perceptron. 
\begin{table}[h]
    \centering
    \scalebox{.75}{
    \begin{tabular}{c|c|c|c|c|}
         Model &  Parameters (M) & FLOPs (G) & I.T. (ms/iter) & T.T (ms/iter)\\
         \hline
         FilterFormer & 0.6 & 11.4 & 39.2 & 136.5\\ \hline
         PatchTST & 3.7 &73.9 & 100.3& 364.8\\ \hline 
         iTransformer & 4.8 & 29.9 & 12.8 & 40.7  \\ \hline
         iFilterFormer & 1.2 & 8.9 & 7.4 & 25.7\\ \hline
         Leddam & 8.5 & 25.24 & 18.6 & 88.5\\ \hline FilterLeddam 
         & 8.5 & 25.25 & 18.6 & 87.2\\ \hline FredFormer
         &  12.1 & 38.32 & 25.2 & 101.5  \\ \hline
    \end{tabular}}
    \caption{Number of trainable parameters in the various models found in table\ref{tab:model_performance_best_scores} for the ECL input length 96 prediction horizon 96 forecasting task. The batch size was fixed at 16.}
    \label{tab:parameter count}
\end{table}

By choosing to remove the MLP from our spectral block, we have that adding the filter blocks contributes a minor number of parameters compared to the transformer architecture.  In Figure 
 \ref{parameter list} we show the number of parameters of our models as a function of the embedding dimension. In practice we only need an embedding dimension of 128-256 and around a total of 4 layers in our model, so that the results in Table \ref{tab:model_performance_best_scores} come from models using in between $0.5$ and $3$ million parameters. In Table \ref{tab:parameter count}, we show the number of parameters needed by our models versus baselines to obtain the results found above. Our models are smaller because we can use a smaller embedding dimension compared to baseline while still obtaining good results. In this regard, while a filter adds more parameters to a given architecture, it can result in a lighter and faster model overall. 
\subsubsection{The Effect of Frequency Filters}
We examine the frequency filters learned by our three models on the electricity dataset for a prediction horizon of 720.

\begin{figure}[h]
    \centering
    \begin{subfigure}[h]{0.8\textwidth}
        \centering
        \includegraphics[width=.75\textwidth]{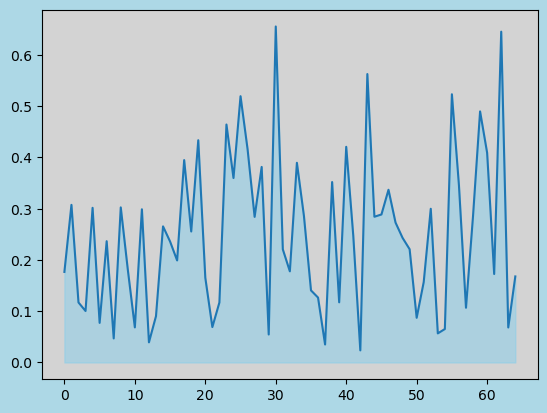} % Replace with your first image
        \caption{FilterFormer} % Fill in the MSE value
        \label{fig:filter_prediction2}
    \end{subfigure}
 
    \begin{subfigure}[h]{0.8\textwidth}
        \centering
        \includegraphics[width=.75\textwidth]{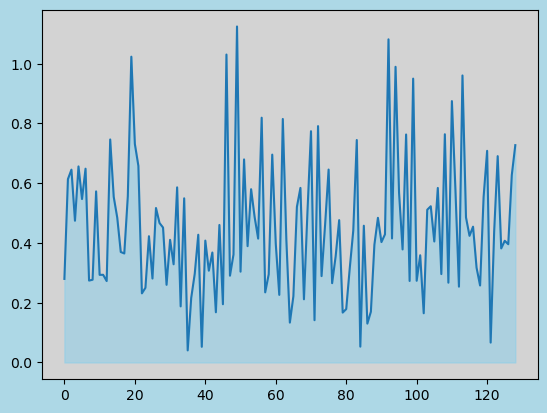} % Replace with your second image
        \caption{iFilterFormer} % Fill in the MSE value
        \label{fig:ifilterform_prediction}
    \end{subfigure}

    \begin{subfigure}[h]{0.8\textwidth}
        \centering
        \includegraphics[width=.75\textwidth]{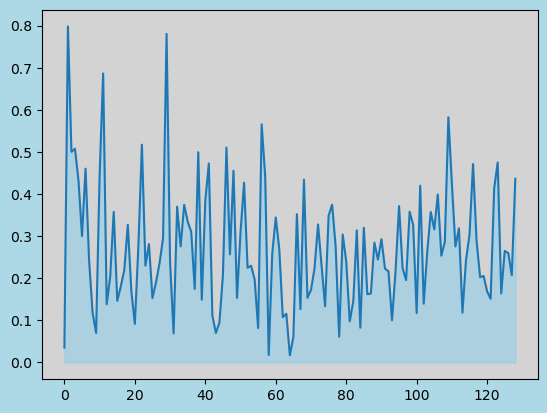} % Replace with your second image
        \caption{FilterLeddam} % Fill in the MSE value
        \label{fig:filterleddam_prediction}
    \end{subfigure}
    \caption{Amplitude spectrums of the filters of our three models on the input 96, prediction horizon 720 task on the Electricity dataset.}

    \label{fig:spectrums}
\end{figure}
There is not a one-to-one correlation between the filters learned by each model. This is in part because our filters are applied to embedded versions of the time series. Thus, if each model settles on a different embedding of the dataset, we would expect different needs from their respective filters. We do note that FilterFormer and iFilterFormer, which are both based upon fairly ``traditional" implementations of self-attention, do have a bias toward middle and high frequency components. More specifically, the most attentuated frequencies occur in the upper two-thirds of the spectrum under consideration by our filter. This is not the case in FilterLeddam, which utilizes different implementations of attention.

\begin{figure}[h]
    \centering
    \includegraphics[width=0.75\linewidth]{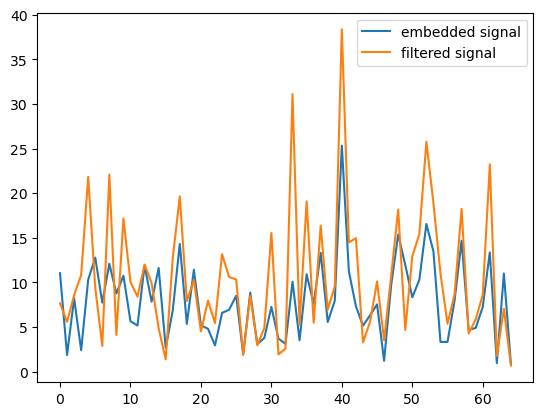}
    \caption{The amplitude spectrum of the embedded signal in our synthetic experiment along with the amplitude spectrum of the embedding after we apply our spectral block.}
    \label{fig:synthetic_spectra}
\end{figure}

To gather insight into how frequency filters enable our forecasting models when we are applying them to embedded signals, we conduct an experiment on a synthetic signal comprised of a low-frequency, mid-frequency, and high-frequency sine wave. In \cite{yi2024filternet} it was noted that the base iTransformer struggles with signals of this nature. This difficulty is often chalked up to the frequency bias of transformers \cite{piao2024fredformer}: theoretical work suggests that self-attention acts as a low-pass filter \cite{shin2024attentive}. Thus, we would expect a transformer-based model to attenuate the high and middle frequencies of its signal as it passes through the attention blocks (we refer to this effect as oversmoothing). It should follow that we test both iTransformer and our iFilterFormer on this signal and plot the results in Figure \ref{fig:signals}. As expected, the model with filter is able to model this signal more effectively. We can analyze the input signal in embedding space before and after the filter is applied to it. As we can see in Figure \ref{fig:synthetic_spectra}, while the filter amplifies many of the frequencies found in the original embedding, the most disproportionally amplified frequencies are those in the mid-high range. In this regard, the filter operates close to a high-pass filter. It follows that filtering can counteract the oversmoothing effect of self-attention, leading to better predictions.
\begin{figure}[h]
    \centering
    \begin{subfigure}[b]{0.8\textwidth}
        \centering
        \includegraphics[width=.7\textwidth]{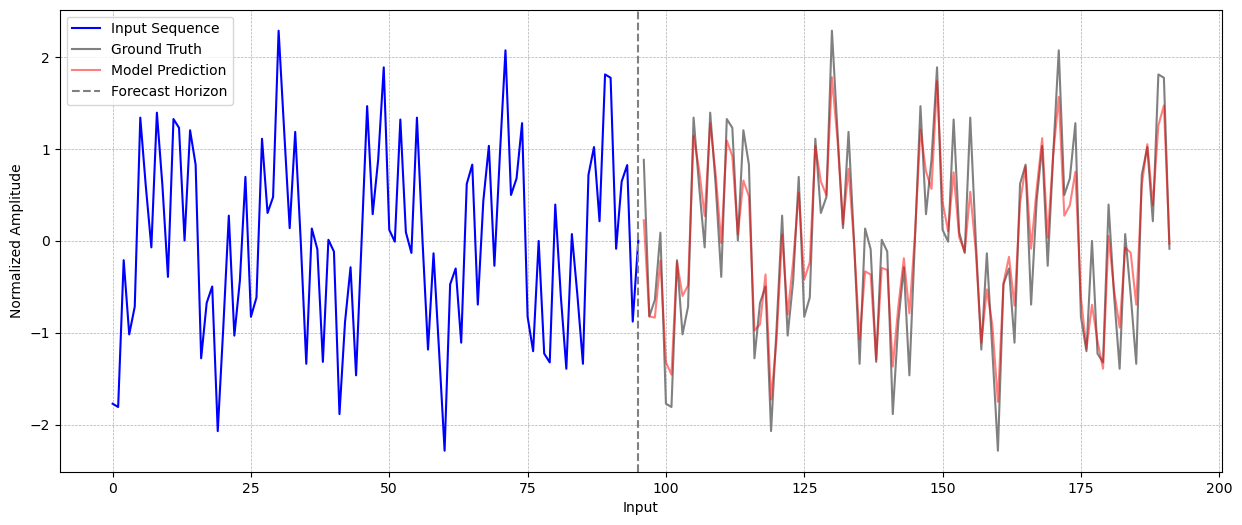} % Replace with your first image
        \caption{MSE = 1.0 e-1} % Fill in the MSE value
        \label{fig:image1}
    \end{subfigure}
    \begin{subfigure}[b]{0.8\textwidth}
        \centering
        \includegraphics[width=.7\textwidth]{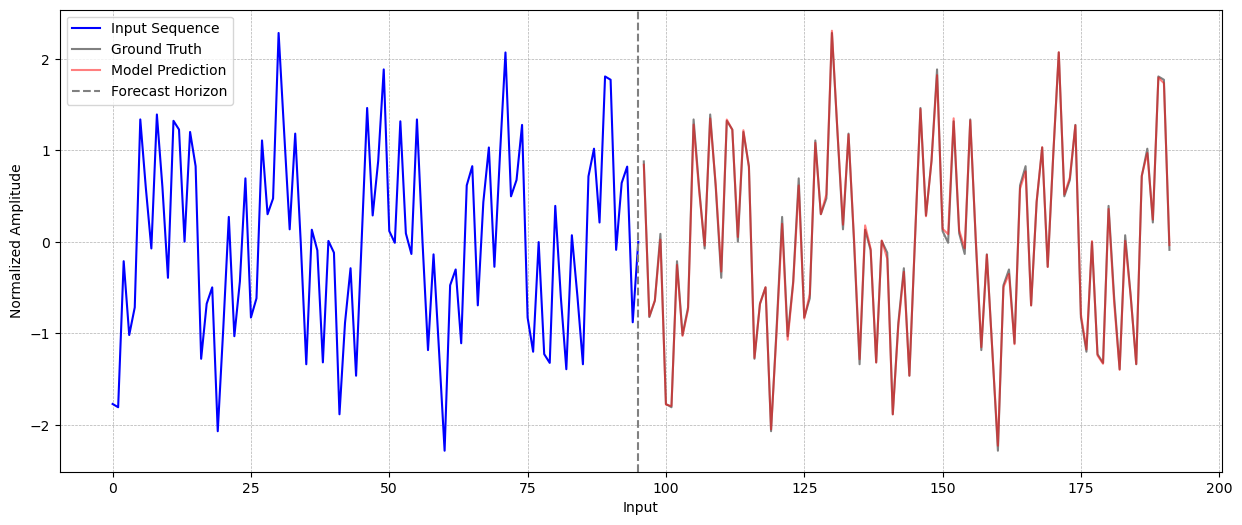} % Replace with your second image
        \caption{MSE = 2.6e-4} % Fill in the MSE value
        \label{fig:image2}
    \end{subfigure}
    \caption{Performance of iTransformer on a simple synthetic signal with low, mid, and high frequencies a) without filter and b) with filter.}
    \label{fig:signals}
\end{figure}

To further validate the impact of our filters, we plot the filters from our FilterFormer model on the ETT and Weather datasets. As with the prior results, the frequency components most amplified by the filters are in the middle to high range.

\begin{figure}[htbp]
    \centering
    \begin{subfigure}[b]{0.48\textwidth}
        \centering
        \includegraphics[width=.7\textwidth]{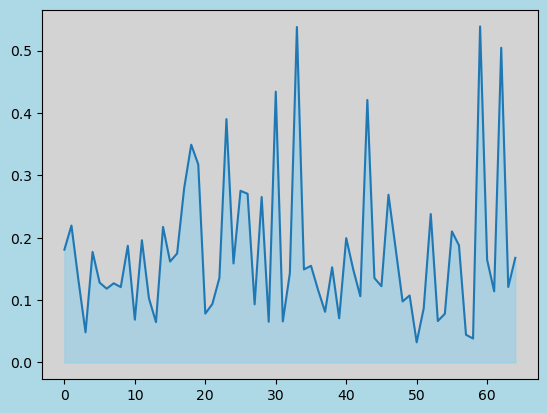} % Replace with your first image
        \caption{ETTm2} % Fill in the MSE value
        \label{fig:ETTm2_Filter1}
    \end{subfigure}
    \begin{subfigure}[b]{0.48\textwidth}
        \centering
        \includegraphics[width=.7\textwidth]{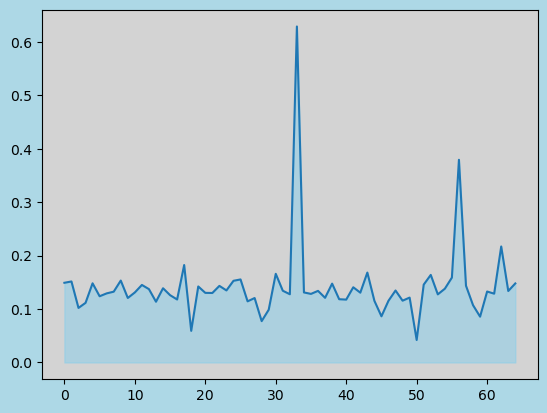} % Replace with your second image
        \caption{Weather} % Fill in the MSE value
        \label{fig:WeatherFilter}
    \end{subfigure}
    \caption{Learned filters from our FilterFormer model on the ETTm2 and Weather datasets. In both tasks the lookback length is 96 and the prediction horizon is 720.}
    \label{fig:Filters1}
\end{figure}

\section{Ablation Studies}

\subsubsection{Varying the Number of Layers} 
\begin{figure}[t]
    \centering
    \includegraphics[width=.75\linewidth]{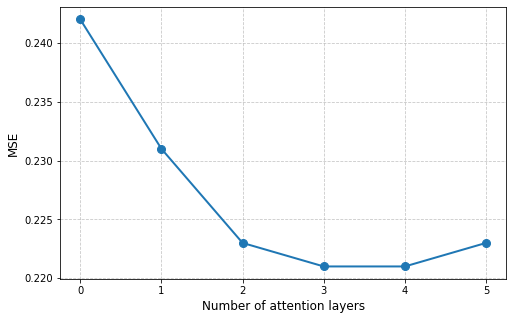}
    \caption{The performance of FilterFormer with 2 spectral gating blocks (with MLP) and a varied number of attention blocks.}
    \label{attention blocks ablation}
\end{figure}

For a fixed number $\alpha$ of filter layers, we are interested in knowing the effect of adding additional attention layers. To study this, we focused on the Electricity dataset with a forecasting horizon of 720. We fixed our model to have two spectral gating blocks, then adjusted the number of attention blocks. The results of this can be seen in figure \ref{attention blocks ablation}. Going from zero attention blocks (as PaiFilter would have) to  multiple attention blocks does lead to a large relative improvement in mean square error. We believe that due to the limited size of our datasets, we quickly see diminishing returns from additional attention blocks to our model.

\subsubsection{Varying the Number of Filters} 
\begin{figure}[h]
    \centering
    \includegraphics[width=0.75\linewidth]{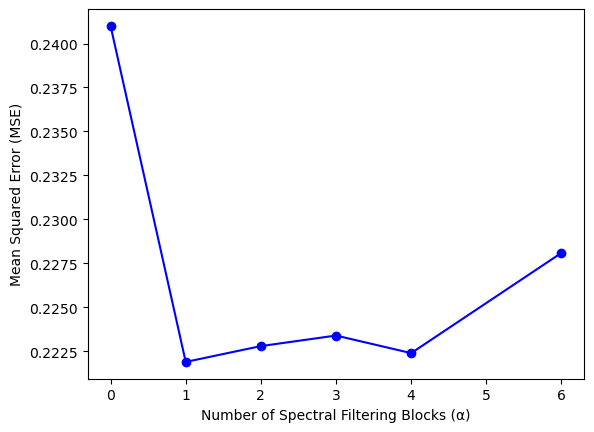}
    \caption{The performance of FilterFormer with 6 layers, $\alpha$ of which are Spectral Blocks and $6-\alpha$ of which are attention blocks}
    \label{fig:filter_ablation}
\end{figure}

Next, we are interested in knowing if there is an optimal balance of spectral blocks versus attention blocks in our FilterFormer model. To investigate this, we train our FilterFormer model on the Electricity dataset with a forecasting horizon of 720. We fix our model to have 6 layers in total, then adjust $\alpha$, the number of spectral layers, from 0 to 6. In this experiment, all of our spectral blocks have a Multilayer Perceptron. We observe in figure \ref{fig:filter_ablation} that while the model works best if there is a mix of filtering and attention, the exact mix is not too important to the forecasting performance. We also note that when the model consists of only spectral blocks, it still outperforms FilterNet. This implies that stacking spectral layers and applying patch embedding are very beneficial for filtered models even if attention is not involved. 

\subsubsection{Placing the Filter before  Patching and Embedding}
One experiment we conducted was whether the learned filter could be placed before the embedding layer of the transformer-based model. If it could, this would be beneficial because it would be even smaller (having only $\mathcal{O}(L)$ parameters, where $L$ is the look-back length) and would be more prone to interpretibility. We thus attempted to move the filter in both FilterFormer and iFilterFormer to occur right after the Reversible Instance Normalization \cite{kim2021reversible} occurs. When we trained these models on the electricity dataset for a prediction horizon of 720, we found that the models performed much worse than when the filters were applied to the embedded signal. 

\begin{table}[h]
    \centering
    \scalebox{.75}{
\begin{tabular}{ll|cc|cc}
\hline
\multicolumn{2}{c|}{Models} & \multicolumn{2}{c|}{FilterFormer} & \multicolumn{2}{c|}{iFilterFormer}\\
 MSE & & .221 & .233 & .208 & .225\\
\hline
\end{tabular}
}
    \caption{Performance of FilterFormer and iFilterFormer on the electricity dataset, prediction horizon of 720. First column indicates results when filter is applied to embedding signal while second column indicates results when filter is applied to original signal.}
    \label{tab:filter_test}
    
\end{table}

\begin{figure}[h]
    \centering
    \begin{subfigure}[b]{0.48\textwidth}
        \centering
        \includegraphics[width=.7\textwidth]{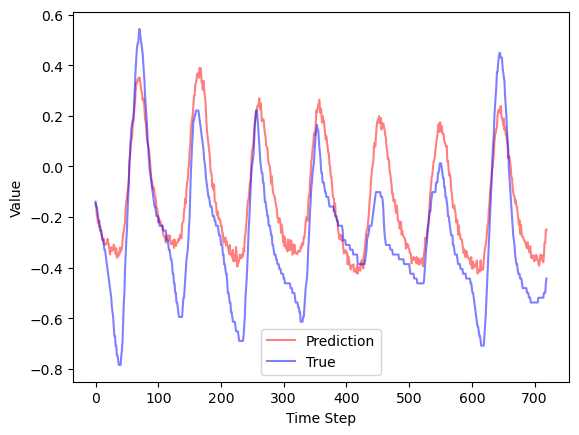} % Replace with your first image
        \caption{ETTm2} % Fill in the MSE value
        \label{fig:ETTm2_prediction}
    \end{subfigure}
    \begin{subfigure}[b]{0.48\textwidth}
        \centering
        \includegraphics[width=.7\textwidth]{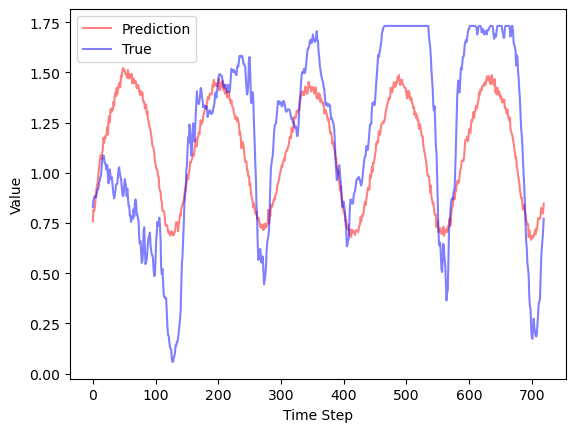} % Replace with your second image
        \caption{Weather} % Fill in the MSE value
        \label{fig:WeatherPrediction}
    \end{subfigure}
    \caption{Predictions from our FilterFormer model with lookback length 96 and prediction horizon 720.}
    \label{fig:Filters2}
\end{figure}

%\subsubsection{Varying the Number of Layers} For a fixed number $\alpha$ of filter layers, we are interested in knowing the effect of adding additional attention layers. To study this, we focused on the Electricity dataset with a forecasting horizon of 720. We fixed our model to have two spectral gating blocks, then adjusted the number of attention blocks. The results of this can be seen in figure \ref{attention blocks ablation}. Going from zero attention blocks (as PaiFilter would have) to  multiple attention blocks does lead to a large relative improvement in mean square error. We believe that due to the limited size of our datasets, we quickly see diminishing returns from additional attention blocks to our model. 

%\begin{figure}
%    \centering
%    \includegraphics[width=.75\linewidth]{images/attention_ablation.png}
%    \caption{The performance of FilterFormer with 2 spectral gating blocks and a varied number of attention blocks.}
%    \label{attention blocks ablation}
%\end{figure}

\section{Conclusions}
In this paper, we demonstrated that adding a learnable filter improves the ability of transformer-based LTSF models. By incorporating the most basic implementation of a learnable filter into popular transformer LTSF models, we were able to create new models that were greater than the sum of their parts while still being lightweight and quick. We validated this effectiveness by testing our new models on datasets spanning a variety of domains and complexities. Even for state of the art forecasting models, we were able to see large improvements in forecasting accuracy with only minor increases in memory and speed requirements. We thus recommend adding a learned filter as a basic configuration for transformer-based forecasting models. It will rarely degrade performance, but it has the potential to drastically improve it. We intentionally stuck to a simple type of filter to demonstrate the efficacy of the method. We believe that further improvements can be made by developing more advanced learnable filters. Along this line, we believe that more work should be done to develop learned filters that are easier to interpret and that are steerable (e.g. a learnable filter that be pushed to be high-pass in nature). Overall, we hope that this work leads to further use of filters in conjunction with transformers in the field of long-term forecasting. 

\section{Acknowledgments} 
The work was partially supported by NSF grants DMS-2151235 and DMS-2219904.

\clearpage
\bibliographystyle{unsrtnat}
\bibliography{main}
%%% Uncomment this line and comment out the ``thebibliography'' section below to use the external .bib file (using bibtex) .




% 	\bibitem{keshet2016prediction}
% 	Keshet, Renato, Alina Maor, and George Kour.
% 	\newblock Prediction-Based, Prioritized Market-Share Insight Extraction.
% 	\newblock In {\em Advanced Data Mining and Applications (ADMA), 2016 12th International 
%                       Conference of}, pages 81--94,2016.

% \end{thebibliography}

\end{document}